\documentclass[conference]{IEEEtran}
\IEEEoverridecommandlockouts
\usepackage[letterpaper, top=0.76in, bottom=1.05in, left=1.01in, right=1.01in]{geometry}

\usepackage{xurl}
\usepackage{cite}
\usepackage{amsmath,amssymb,amsfonts}
\usepackage{algorithmic}
\usepackage{graphicx}
\usepackage{textcomp}
\usepackage{xcolor}
\usepackage{mathtools}
\usepackage{float}
\usepackage{dirtytalk}
\usepackage{algorithm}
\usepackage{subfigure}
\usepackage{hyperref}
\usepackage{amsmath}

\begin{document}
\title{Channel Balance Interpolation in the Lightning Network via Machine Learning\\
\thanks{* Jointly supervised }
}
% \author{\IEEEauthorblockN{ Anonymized for Double-Blind Review}}
\author{\IEEEauthorblockN{Vincent Davis}
\IEEEauthorblockA{
\textit{Amboss Technologies}\\
Nashville, USA \\
v@amboss.tech}
\and
\IEEEauthorblockN{ Vikash Singh*}
\IEEEauthorblockA{\textit{Stillmark} \\
San Francisco, USA \\
vikash@stillmark.com}
\and
\IEEEauthorblockN{ Emanuele Rossi*}
\IEEEauthorblockA{\textit{Amboss Technologies} and \textit{VantAI} \\
Barcelona, Spain \\
emanuele.rossi1909@gmail.com}
}

\maketitle

\begin{abstract}
The Bitcoin Lightning Network is a Layer 2 payment protocol that addresses Bitcoin's scalability by facilitating quick and cost-effective transactions through payment channels. This research explores the use of machine learning models to interpolate channel balances within the network, which can be used for optimizing the network's pathfinding algorithms. While there has been much exploration in balance probing and multipath payment protocols, predicting channel balances using solely node and channel features remains an uncharted area. This paper evaluates the performance of several machine learning models against two heuristic baselines and investigates the predictive capabilities of various features. Our model performs favorably in experimental evaluation, reducing the relative error by 27\% compared to an equal split baseline where both edges are assigned half of the channel capacity.
\end{abstract}


\section{Introduction}
The Lightning Network is a layer-2 protocol on the Bitcoin blockchain that enables rapid and cost-effective payments. Central to optimizing its efficiency is the accurate estimation of channel balances, which directly informs pathfinding strategies. Effective pathfinding circumvents the need for a cumbersome trial-and-error process in identifying viable payment routes. While there have been advances in understanding pathfinding with uncertain channel balances \cite{pickhardt2021security}, exploration of reinforcement learning (RL) for pathfinding \cite{valko2023increasing} and the development of multipath payment protocols \cite{pickhardt2021optimally}, the direct prediction of channel balances is less explored territory.

This gap in the literature motivates the current study, which seeks to determine whether it is feasible to interpolate the balances of channels within the Lightning Network accurately. Furthermore, this work aims to identify the most predictive features for balance interpolation, questioning whether these predictions can rely solely on node features, channel features, or a combination of both, and possibly enhanced by topological information of the network. To tackle these questions, the study evaluates the performance of two baseline methods against six machine learning (ML) models, each with a varying subset of features.

The investigation reveals that ML models can predict channel balances more accurately than heuristic methods like the equal split assumption. This predictive power can be leveraged to refine pathfinding algorithms by reducing uncertainty in channel balances and decreasing failed payment attempts.

% This study sheds light on the potential of specific features and how machine learning can enhance the network's pathfinding efficiency.

% The remainder of this paper is organized as follows. \hyperref[sec:background]{Section II} provides background information on Bitcoin and the Lightning Network. \hyperref[sec:motivation]{Section III} elaborates on the specific problem and the motivation for improved channel balance interpolation. \hyperref[sec:relatedwork]{Section IV} reviews related work in areas such as pathfinding algorithms and multipath payment protocols for the Lightning Network. Sections \hyperref[sec:problem]{V},\hyperref[sec:data]{VI}, and \hyperref[sec:mlmodels]{VII} formally define the problem statement, the data collection and preprocessing steps, and the machine learning models and features evaluated. \hyperref[sec:evaluation]{Section VIII} outlines the methodology and metrics. Sections \hyperref[sec:results]{IX} and \hyperref[sec:futurework]{X} present and analyze the performance of the various models. Finally, \hyperref[sec:discussion]{Section XI} explores potential future work, such as implementing an enhanced pathfinding algorithm that leverages the balance interpolation model and directions for further improving model performance.

\section{Background}
\label{sec:background}
\subsection{Bitcoin}
Bitcoin is a peer-to-peer electronic cash system invented in 2008 by a person or group under the pseudonym Satoshi Nakamoto \cite{Satoshi}. The purpose of Bitcoin is to provide a trustless way to transfer value. The features which allow this trustless transfer of value are Bitcoin's distributed ledger, Proof-of-Work consensus model, and fixed supply. The Bitcoin network operates without relying on a central server. New transactions are added to the decentralized ledger through a process known as mining. Bitcoin nodes which participate in this mining process collect transactions and add them to to an \say{ongoing chain of hash-based proof-of-work} \cite{Satoshi}, by solving a computationally difficult problem. The chain which demonstrates the most computational effort is regarded as the authentic record of transactions. 

%Proof-of-Work adds a real world cost to attempting to invalidate past transactions, also known as a block re-organization.  

Transaction fees in Bitcoin are determined by the amount of data consumed on-chain rather than the value being transferred. All else equal, this pricing dynamic makes smaller payments more expensive relative to larger payments. Additionally, the Bitcoin blockchain's transaction throughput is intentionally constrained. With a fixed block size and block time, the network is designed to process an average of one 4MB block every ten minutes, translating to a maximum of seven transactions per second \cite{scaling}. While this design choice ensures accessible storage and bandwidth requirements for Bitcoin, it also hampers the scalability of its blockchain as a peer-to-peer electronic cash system. Considering the time, fees, and energy expended, common use cases like micropayments, subscriptions, and streaming payments are prohibitively expensive to execute on the base layer of the Bitcoin blockchain.

\subsection{Lightning Network}
The Lightning Network is a Layer 2 payment protocol built on top of the Bitcoin blockchain \cite{Poon2016}. LN emerged in 2017 as a potential solution to Bitcoin's scalability challenge that preserves its trustless nature. In LN, value is transferred by way of payment channels between peers in the network. A payment channel is established by depositing funds into a 2-of-2 multisig address jointly controlled by both peers. This deposit is known as the funding transaction. The funding transaction establishes the initial balance of funds between both parties. 

In contrast, the latest balance of the channel is contained in the commitment transaction. Both peers maintain a copy of this transaction but defer its broadcast until they want to close the channel. When payments are made through this channel, the peers jointly update the commitment transaction. Because the peers defer broadcasting the commitment transaction, payments on the Lightning Network are not subject to the 10-minute average blocktime on Layer 1 (the Bitcoin blockchain). Furthermore, the pricing dynamic of LN is fundamentally different from that of L1. While transactions on L1 compete for blockspace by paying higher fees to miners, routing nodes on LN compete with each other to relay payments by offering lower fees.

\begin{figure}
    \begin{subfigure}
    \centering
    \includegraphics[width=0.95\linewidth]{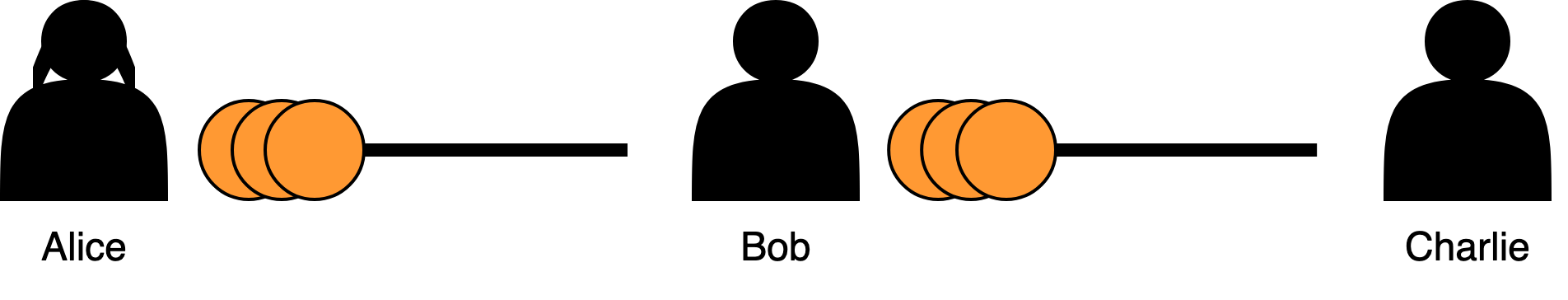}
    \caption{Single path payments can only occur if there is sufficient liquidity. In this network state, Bob can relay a payment from Alice to Charlie, but he cannot relay a payment from Charlie to Alice.}
    \label{fig:abc}
\end{subfigure}
% \\
\vspace{5pt}
\begin{subfigure}
    \centering
    \includegraphics[width=0.95\linewidth]{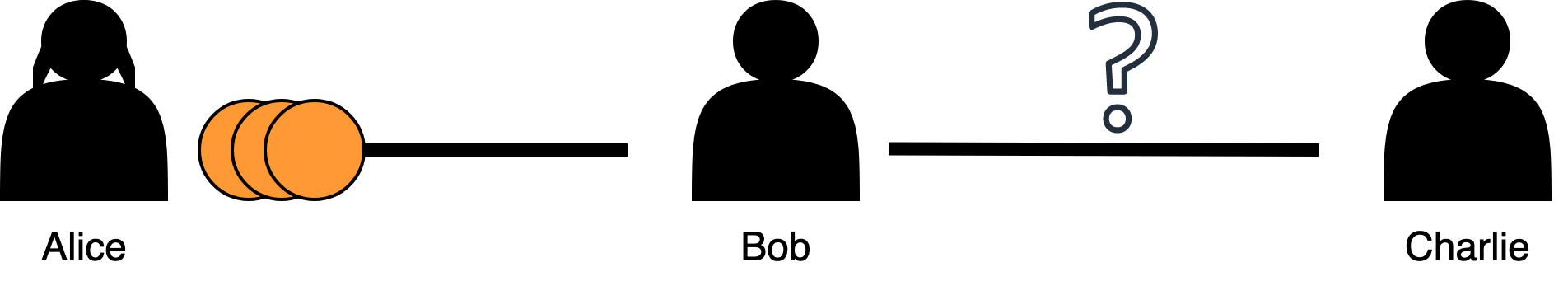}
    \caption{Alice only knows the balances of channels she owns. She may learn information about intermediate channels after attempting payments.}
    \label{fig:def}
\end{subfigure}
\end{figure}

Of course, there are trade-offs to using LN over L1. Single-path payments can only be relayed if there exists a path with sufficient liquidity from source to destination. In other words, one cannot forward a payment if they have an insufficient balance on their side of the channel. However, only the nodes incident to a channel know the balance of that channel. An example of these concepts can be seen in \autoref{fig:abc} and \autoref{fig:def}. Thus, payment pathfinding in the Lightning Network is a trial-and-error process.

\section{Motivation}
\label{sec:motivation}
Topological analysis from other works has shown that the Lightning Network topology is consistent with the small-world architecture \cite{seres2020topological}. These graphs exhibit robustness because of their high local clustering coefficient and low average path length. However, in the Lightning Network, this abundance of short paths comes with the downside of not knowing which ones have sufficient liquidity to relay the payment. For example, River is a Bitcoin-only exchange that operates two of the largest nodes in the Lightning Network \cite{RF1, RF2}. In a research report by River published in 2023, they cited timeout during pathfinding as the leading cause of payment failure for their nodes \cite{riverfinancial}. The purpose of this work is to investigate ways of evaluating paths \textit{before} trying them. Our key focus is on whether the balance of intermediate channels can be interpolated from channel policies and node metrics. This would enable the prioritization of paths in the trial-and-error process and save time in the search for a viable path. 

\section{Related Work}
\label{sec:relatedwork}

\textit{LightningNetworkDaemon}, the majority implementation present on the Lightning Network \cite{zabka2022empirical}, has two models for estimating the success probabilities of paths: \textit{apriori} and \textit{bimodal}. The \textit{apriori} model assumes a constant probability on untested channels, taking into account channel capacity and payment size \cite{lightningnetworkdaemon}.
The \textit{bimodal} model  is an outcome of Pickhardt et al.'s work \cite{pickhardt2021security} where they sampled balances in the Lightning Network and found that portions of channels followed different distributions. They observed that about 30\% of the channel balances were concentrated on one side while the rest of the channel balances made up a uniform distribution. The assumption of a bimodal distribution is a configurable path estimator option in LND as of version v0.16.

The next majority implementation present on the Lightning Network is \textit{core-lightning}. In the default pathfinding behavior, paths are weighted using a fee and risk factor. This weight takes into account the risk of locking up liquidity for some number of blocks, with bias to larger capacity channels. The \textit{core-lightning} implementation also includes a plugin architecture to extend and customize its functionality. One such plugin, altpay, implements pickhardt payments with probability scoring. This payment algorithm is the result of work by Pickhardt and Richter to generalize payments as a minimum-cost flow problem \cite{pickhardt2021optimally}. Here, the cost considers both the fee and the reliability of a path. The reliability of the path is estimated using an uncertainty network.  The uncertainty network is updated in a similar fashion to the trial-and-error approach in a single-path payment. However, multipath payments explore multiple paths, are split into smaller amounts, and are not constrained by the liquidity of a single route. The authors found that multipath payments move funds across the network more reliably than single-path payments.

The work of Tikhomirov et. al demonstrates how one can systematically probe channels by leveraging payment errors and response times, effectively revealing balances \cite{probing}. However, these methods impose network overhead, require capital, and compromise privacy. The reliance on active interaction makes probing costly and intrusive, as it requires continuous test transactions to infer balances. The capital requirement is two-fold: not only does probing demand upfront liquidity, but it also limits the recoverable balances to amounts smaller than the deployed capital.

Other works include Valko and Kudenko's work on using reinforcement learning for path planning in the Lightning Network \cite{valko2023increasing}. In this technique, an agent learns to predict paths in the Lightning Network that have low total cost and are close to the shortest path calculated by Dijkstra's algorithm. However, the paper does not explicitly consider real-world channel balances in the Lightning Network.

\section{Problem Formulation}
\label{sec:problem}
The Lightning Network can be modeled as a directed graph \( G = (V, E) \), where \( V \) represents the set of nodes and \( E \) represents the set of edges. Each node \( u \) and each edge \( (u, v) \) have associated features, denoted by \( \mathbf{x}_u \in \mathbb{R}^k \) for nodes and \( \mathbf{e}_{(u, v)} \) for edges, which contain specific information about those nodes and edges. Additionally, each edge \( (u, v) \) has a scalar value \( y_{(u, v)} \geq 0 \), representing the pre-allocated balance for transactions from \( u \) to \( v \).

Graph \( G \) has the constraint that if an edge exists in one direction, it must also exist in the opposite direction, i.e., \( (u, v) \in E \iff (v, u) \in E \). The set of two edges between a pair of nodes is called a channel, denoted as \( \{(u, v), (v, u)\} \). For simplicity, we represent a channel by the set of its two nodes: \( \{u, v\} \). The total capacity of the channel \( \{u, v\} \) is defined as \( c_{\{u, v\}} = y_{(u, v)} + y_{(v, u)} \).

We are provided with the total channel capacities \( c_{\{u, v\}} \) for all channels in the graph, but we only know the individual values \( y_{(u, v)} \) for a subset of edges. Note that knowing \( y_{(u, v)} \) allows us to determine \( y_{(v, u)} \), since \( y_{(v, u)} = c_{\{u, v\}} - y_{(u, v)} \). Therefore, we can focus on predicting \( y_{(u, v)} \).

Moreover, since we are given \( c_{\{u,v\}} \) for all edges and we know \( 0 \leq y_{(u, v)} \leq c_{\{u,v\}} \), we also know that \( y_{(u, v)} = p_{(u,v)}c_{\{u,v\}} \), where \( 0 \leq p_{(u,v)} \leq 1 \). Intuitively, \( p_{(u,v)} \) is the proportion of the channel capacity that belongs to the \( (u,v) \) direction. From this, we focus on predicting \( p_{(u,v)} \), and then use it to obtain \( y_{(u, v)} \).

Therefore, our primary task is to predict \( p_{(u,v)} \) for all edges where it is unobserved.

\section{Data Collection and Preprocessing}
\label{sec:data}
The data used in this experiment is a combination of publicly available information from the Lightning Network and crowdsourced information from nodes in the network. 
This experiment uses a network snapshot from December 15, 2023. The snapshot was collected from a lightning node on the network operated by Amboss Technologies. Further information is derived from the network snapshot. This information includes aggregated channel capacities, giving the capacity of each node and the network. Other relevant information includes the fee ratios of each node which is found by dividing average incoming fee rates by average outgoing fee rates. In other works \cite{bitmex}, this ratio is referred to as drain. Balance data is crowdsourced from nodes in the Lightning Network. Node operators may opt to share balance information with Amboss Technologies in order to receive liquidity notifications, automatically purchase channels, or maintain a historical account.  
\subsection{Preprocessing Steps}
Each node's balance is represented by its local balance, recorded at one-minute intervals over the past hour. This data is converted into a probability density function (PDF) through kernel density estimation. Subsequently, a balance value is sampled from each PDF, serving as the representative local balance for the corresponding channel within the dataset.

\subsection{Preliminary Analysis}

The following section is a statistical overview of the Lightning Network, including various tables and metrics that detail the distribution of channel capacities and the ratio of channel balances to their capacities. The analysis of channel capacities highlights the range and distribution of liquidity across the network. This data serves as a reference for understanding the current state and structural characteristics of the Lightning Network.

\begin{figure*}
    \begin{subfigure}
        \centering
        \includegraphics[width=0.3\linewidth]{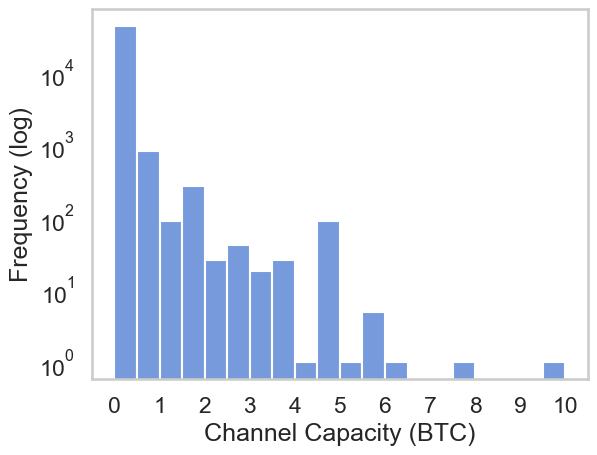}
        \label{fig:dist-channels}
    \end{subfigure}
    \begin{subfigure}
        \centering
        \includegraphics[width=0.3\linewidth]{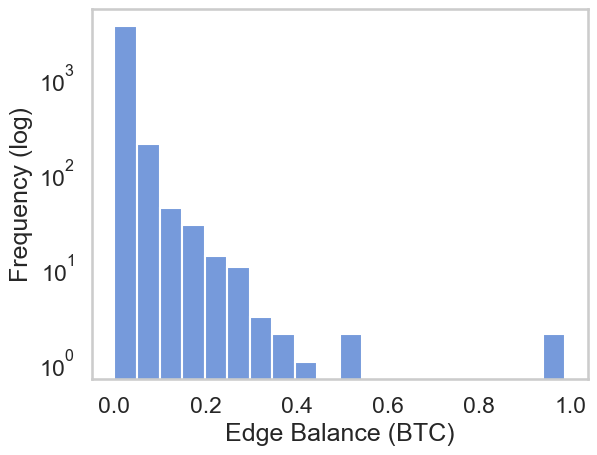}
        \label{fig:dist-edges}
    \end{subfigure}
    \begin{subfigure}
        \centering
        \includegraphics[width=0.3\linewidth]{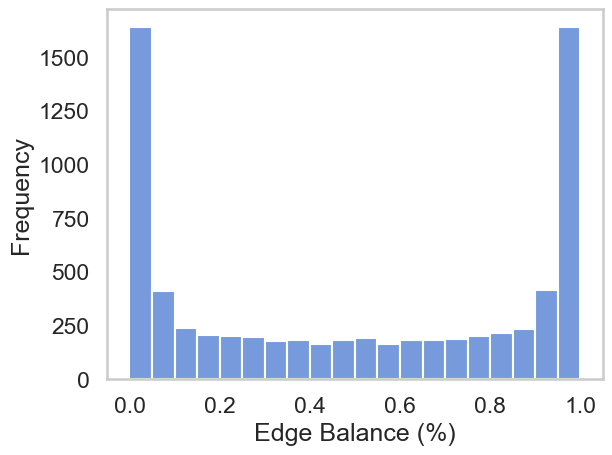}
        \label{fig:bimodal}
    \end{subfigure}
\caption{Distribution of channel capacities, crowdsourced edge balances, and normalized edge balances in the Lightning Network. }
        \label{fig:allthree}
\end{figure*}

The leftmost histogram in \autoref{fig:allthree} shows the distribution of channel capacities in the Bitcoin network. Most channels have capacities between 0 and 10 BTC, with the highest concentration at the lowest capacity range. Frequencies decline sharply for larger capacities, though distinct peaks suggest a preference for round-number values. Overall, smaller channel capacities dominate, while larger ones are relatively rare.

The middle histogram in \autoref{fig:allthree} depicts the distribution of known edge balances denominated in Bitcoin (BTC). This data also shows a left-skewed distribution, indicating a prevalence of smaller balances over larger ones. 

The rightmost histogram in \autoref{fig:allthree} depicts the distribution of known edge balances as a percentage of their respective capacities. It reveals a bimodal pattern, with peaks at balances near empty and near full capacity, while intermediate balances are notably scarce. This suggests that channels are typically either minimally used or nearly fully utilized, rather than maintaining mid-range balances. These findings align with those reported by Pickhardt et al. \cite{pickhardt2021security}.

\subsection{Feature Analysis}

To evaluate the relationship between each feature and channel balance, a correlation analysis was performed. \autoref{fig:correlation-table} presents features with notable correlations, defined as having a correlation coefficient outside the range of \([-0.1, +0.1]\). The \textbf{Max HTLC} feature represents the maximum outgoing payment size (denominated in millisatoshis) a node will process. \textbf{Feature flag \#19} is a binary indicator of whether a node allows channels exceeding approximately 16.7 million satoshis, commonly known as the \say{wumbo channel} threshold in the Lightning Network. \textbf{Local} indicates that the feature pertains to the outgoing edge, while \textbf{remote} refers to a feature of an incoming edge.

\begin{figure}[H]
    \begin{table}[H]
        \caption{Features and their correlations to channel balance.}
            \label{fig:correlation-table}
        \centering

% \resizebox{0.5\linewidth}{!}{%
\begin{tabular}{ll}
Feature                & Correlation \\ \hline
Local Maximum HTLC  & 0.6246    \\ 
Remote Maximum HTLC  & 0.6165    \\ 
Remote Feature 19 is Known  & 0.1603    \\ 
Local Feature 19 is Known  & 0.1523   \\ 
\end{tabular}%
% }
\end{table}
\end{figure}

Both Maximum HTLC and Feature 19 show a positive correlation with the target variable, as indicated by their positive correlation coefficients. Local Maximum HTLC and Remote Maximum HTLC exhibit a notably strong positive correlation. In contrast, Local Feature 19 is Known and Remote Feature 19 is Known exhibit a moderate yet meaningful positive correlation.

\section{Methodology}
\label{sec:mlmodels}
\subsection{Modeling}
 We will predict $p_{(u,v)}$ by learning a parametric function:
 \[
\hat p_{(u,v)}= f_{\Theta}(u, v, G, \mathbf{x}_u, \mathbf{x}_v, \mathbf{e}_{(u, v)}, c_{\{u, v\}})
 \]
where $\Theta$ are learnable weights, $\mathbf{x}_u$, $\mathbf{x}_v$ and  $\mathbf{e}_{(u, v)}$ are the node and edge features respectively, while $c_{\{u, v\}}$ is the channel capacity.
While several choices are possible for $f_{\Theta}$, such as multi-layer perceptrons or Graph Neural Networks, we focus on Random Forests for this work given their simplicity and efficacy. In particular, our Random Forest (RF) model operates on the concatenation of the features of the source and destination nodes as well as the edge features:
\[
\hat p_{(u,v)}=\mathrm{RF}(x_u || z_u || x_v|| z_v || e_{(u,v)})
\]

The model is trained using a Mean Squared Error loss. Moreover, we know that $p_{(u,v)}=1-p_{(v,u)}$. Therefore, we would like to design our model such that $f_{\Theta}(u, v, G, x, c)=1-f_{\Theta}(v, u, G, x, c)$. We can obtain this approximately through data augmentation: make sure that both directions of an edge are present in the data so that the model is trained to predict both $f_{\Theta}(u, v, G, x, c)=p_{(u,v)}$ and $f_{\Theta}(v, u, G, x, c)=1-p_{(v,u)}$.

\subsection{Node Features}

\begin{itemize}
    \item \textbf{Node Feature Flags}\\
0-1 vector indicating which of the features each node supports. For example, feature flag \#19, the wumbo flag.
    \item \textbf{Capacity Centrality}\\
The ratio of a node's capacity to the total network capacity, indicating the proportion of network capacity connected to the node.
\item \textbf{Fee Ratio}\\
The ratio of a node's average outgoing fee to its average incoming fee.
\end{itemize}

\subsection{Edge Features}
\begin{itemize}
    \item \textbf{Time Lock Delta} 
The number of blocks a relayed payment is locked into an HTLC.
    \item \textbf{Min HTLC} 
The minimum amount this edge will route. (denominated in millisatoshis)
    \item \textbf{Max HTLC} 
The maximum amount this edge will route. (denominated in millisatoshis)
    \item \textbf{Fee Rate} 
Proportional fee to route a payment along an edge, measured in parts per million (ppm) satoshis. (1 ppm = 1 satoshi earned for every million satoshi routed).
    \item \textbf{Fee Base} 
Fixed fee to route a payment along an edge. (denominated in millisatoshis)
\end{itemize}

\subsection{Positional Encodings}
Positional encoding is essential for capturing the structural context of nodes, as graphs lack inherent sequential order. Using eigenvectors of the graph Laplacian matrix as positional encodings provides a robust solution to this challenge. These eigenvectors highlight key structural patterns about the overall topology of the graph \cite{vijay}. By incorporating these spectral properties, machine learning models can better identify and leverage network-wide characteristics, improving performance in tasks such as node classification and community detection.

In practice, computing positional encodings for networks of the scale studied in this work ($7,497$ nodes and $40,462$ edges) takes approximately one second on a standard machine. While this computation could become more expensive on significantly larger graphs, the current performance is sufficient for near realtime use cases on today’s Lightning Network.

\section{Model Evaluation}
\label{sec:evaluation}
We set aside $10\%$ of the observed $y_{(u,v)}$ as our test set and $10\%$ as our validation set. 
We use Mean Absolute Error (MAE) to evaluate how well our predictions match the test set. In particular, we consider two MAE metrics,
$MAE_y$ which is computed on the raw channel balance, and $MAE_p$, computed on the channel proportion on capacity.
\[
MAE_y=MAE(y, \hat y)=\sum_{(u,v) \in test} \left| y_{(u,v)}-\hat y_{(u,v)} \right| \]
\[
MAE_p=MAE(p, \hat p)=\sum_{(u,v)\in test} \left|p_{(u,v)}-\hat p_{(u,v)}\right| 
\]

Our primary objective is to minimize the prediction error of channel balances while ensuring the model generalizes well across different network conditions. Specifically, we seek to learn a function \( f_{\Theta} \) that predicts the balance proportion \( \hat p_{(u,v)} \) for each edge \( (u,v) \) using node, edge and network features.

\subsection{Baselines}
\subsubsection{Equal Split}
Split the capacity equally between the two edges: 
\[
\hat p_{(u,v)}= \hat p_{(v,u)}=0.5
\].

\subsubsection{Local Max HTLC}
Use the max HTLC amount from the local channel policy as the balance:
\[
\hat p_{(u,v)}= \mathrm{max\_htlc}(e_{(u,v)})
\]
Note that this value is normalized by the channel's capacity. It is also not subject to the constraint that $\hat p_{(u,v)} + \hat p_{(v,u)} = 1$.

\subsection{ML Models}

\subsubsection{Edge-Wise Random Features Prediction}
Assign random features $r_{(u,v)}$ from an isotropic Normal distribution to each edge.
Train a model that predicts $p_{(u,v)}$ only as a function of the random features of the edge $(u,v)$:
\[
\hat p_{(u,v)}=\mathrm{RF}(r_{(u,v)})
\]
\subsubsection{Node-Wise Prediction}
A model which predicts $p_{(u,v)}$ only as a function of the features of the source node $u$:
\[
\hat p_{(u,v)}=\mathrm{RF}(x_u)
\]

\subsubsection{Edge-Wise Prediction}
A model which predicts $p_{(u,v)}$ only as a function of the features of the edge $(u,v)$:
\[
\hat p_{(u,v)}=\mathrm{RF}(e_{(u, v)})
\]

\subsubsection{Concatenated Prediction}
A model which predicts $p_{(u,v)}$ as a function of the concatenation of the features of the source and destination nodes as well as edge features:
\[
\hat p_{(u,v)}=\mathrm{RF}(x_u || x_v || e_{(u,v)})
\]
\subsubsection{Shallow Graph Prediction}
Compute graph positional encodings (an embedding for each node representing its position in the graph) as graph laplacian eigenvectors. Let’s represent such positional encodings for node $u$ by $z_u$. Train a model that predicts $p_{(u,v)}$ as a function of the concatenation of the positional encodings of the source and destination nodes:
\[
\hat p_{(u,v)}=\mathrm{RF}(z_u || z_v)
\]

\subsubsection{Joint Prediction}
A model which predicts $p_{(u,v)}$ as a function of the concatenation of the node features, edge features and the positional encodings:
\[
\hat p_{(u,v)}=\mathrm{RF}(x_u || x_v || e_{(u,v) || z_u || z_v})
\]

\subsection{Methodology}
Each model was evaluated using a train-test split method. The performance of each model was reported in a table using four metrics:
\begin{itemize}
    \item \( MAE_y \): Mean Absolute Error between actual and predicted edge balances.
    \item \( MAE_p \): Mean Absolute Error in edge balance predictions.
    \item \( R \): Correlation coefficient, measuring the alignment between predictions and actual values.
    \item \( R^2 \): Coefficient of determination, representing the proportion of variance in edge balances explained by the model.
\end{itemize}

% The model could be any kind of machine learning model. In this study, we used a Random Forest regression model. 

This table allows for a straightforward comparison of model performance, focusing on accuracy (\( MAE\)) and the strength of the relationship between predicted and actual values (\( R \) and \( R^2 \)). The goal being to identify the model that offers the best balance between minimizing prediction errors and maximizing explanatory power for edge balances. Additionally, the Mean Decrease Impurity (MDI) feature importance was plotted, highlighting each feature's contribution to the model's effectiveness.

\section{Results}
\label{sec:results}
The results are recorded in Table \ref{fig:performance-summary}. Note that MAE$_y$ is denoted in millions of satoshis.
\begin{figure}[H]
\begin{table}[H]
\caption{Model Performance}
    \label{fig:performance-summary}
\resizebox{\linewidth}{!}{%
\begin{tabular}{lllll}
Model          & MAE$_p$($\downarrow$) & MAE$_y$($\downarrow$) & R ($\uparrow$)     & $R^{2}$ ($\uparrow$) \\ \hline
Equal Split    & $0.358$  & $1.60$   & $-$     & $-0.001$              \\
Local Max HTLC & $0.479$  & $2.16$   & $0.034$ & $0.001$                \\
Random Edge Features         & $0.362$  & $1.67$   & $-0.06$ & $-0.03$       \\
Nodes          & $0.313$  & $1.31$   & $0.355$ & $0.097$                \\
Edges          & $0.316$  & $1.39$   & $0.350$ & $0.115$                \\
Concatenated   & $0.265$  & $1.09$   & $0.601$ & $0.351$                \\
Shallow        & $0.285$  & $1.20$   & $0.530$ & $0.275$                \\
$\textbf{Joint}$          & $\textbf{0.259}$  & $\textbf{1.08}$   & $\textbf{0.612}$ & $\textbf{0.365}$               
\end{tabular}%
}
\end{table}
    % \caption{Performance summary of all models and heuristics.}
\end{figure}

The Equal Split and Local Max HTLC heuristics provide initial reference points for comparing performance. The Equal Split and Local Max HTLC models demonstrate negligible predictive power, explaining virtually none of the variance in the dependent variable. The Node-wise and Edge-wise models show some predictive ability but still limited effectiveness. Moderate improvements are seen in the Concatenated and Shallow Graph models. The Joint model, which considers node, edge, and positional encoding features, has the greatest predictive power.

The inclusion of positional encodings in the Random Forest model resulted in improved performance compared to models without positional encodings. This suggests that positional encodings enhance the model's ability to capture the underlying data structure. Additionally, it should be noted that the shallow graph model outperformed the Equal Split heuristic, indicating that positional information alone is predictive of channel balances.

\begin{figure*}[]
    \begin{subfigure}
        \centering
        \includegraphics[width=0.31\linewidth]{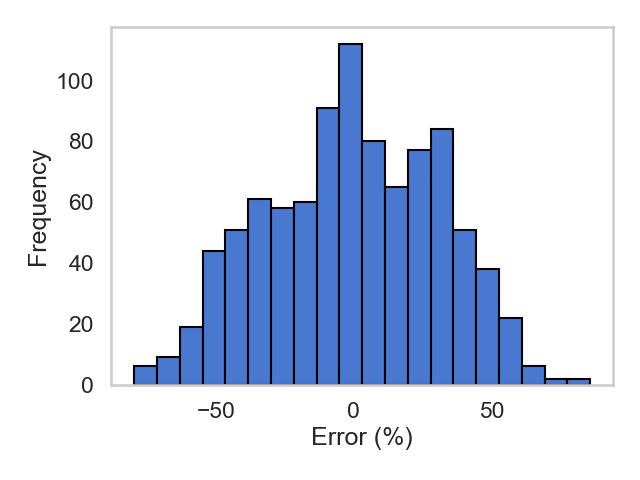}
        \label{fig:pistogram}
    \end{subfigure}
    \begin{subfigure}
        \centering
        \includegraphics[width=0.31\linewidth]{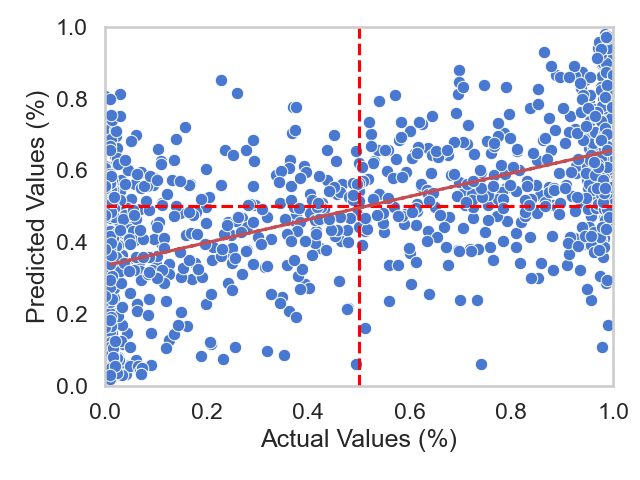}
        \label{fig:pscatterplot}
    \end{subfigure}
    \begin{subfigure}
        \centering
        \includegraphics[width=0.31\linewidth]{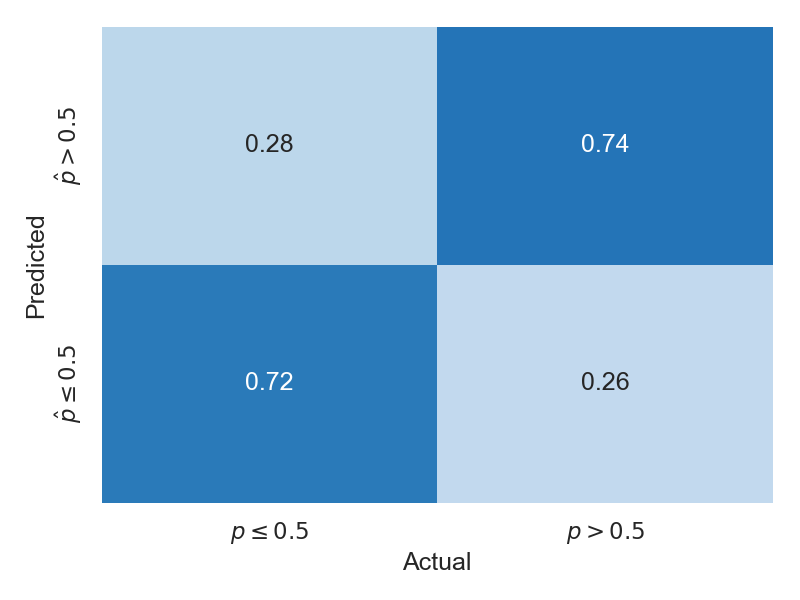}
    \label{fig:pconfusion}
    \end{subfigure}
    \caption{Histogram of errors, a scatterplot of the actual vs. predicted values, and a confusion matrix of the best performing model.}
    \label{fig:threediagrams}
\end{figure*}

The histogram in \autoref{fig:threediagrams} illustrates the error distribution for the top-performing model. Errors are binned between -1 and 1, with the y-axis representing their frequency. The distribution peaks around zero, tapering off symmetrically on both sides, indicating that large errors are less frequent and that the model does not systematically overpredict or underpredict.  

The scatterplot in \autoref{fig:threediagrams} compares actual and predicted balance percentages. The best-fit line has a positive slope, suggesting that predicted values generally increase with actual values, though the correlation is not particularly strong.  

The confusion matrix in \autoref{fig:threediagrams} reveals that the model effectively identifies which side holds the majority of the liquidity. Notably, for test data where there is less local liquidity than remote, the model correctly predicted \( \hat{p} \leq 0.5 \) 72\% of the time. The resulting precision is $0.74$ and the recall is $0.7255$. 

\subsection{Feature Importance}

Feature importance analysis using Mean Decrease in Impurity (MDI) provides a clear comparison of the predictive ability of each feature in the Random Forest model. 

% \begin{figure}[H]
%     \centering
%     \includegraphics[width=\linewidth]{figures/Feature_Importance.png}
% \end{figure}

\begin{table}[h]
    \caption{Feature importance using mean decrease in impurity}
    \centering
    \begin{tabular}{l c}
        \textbf{Feature} & \textbf{Importance (\%)} \\
        \hline
        Max HTLC & 5.0 \\
        Local Fee & 8.3 \\
        Remote Fee & 12.5 \\
        Fee Ratio & 10.2 \\
        Fee Rate (ppm) & 14.7 \\
        Capacity Centrality & 16.0 \\
        Positional Encoding & 33.3 \\
    \end{tabular}
    \label{tab:feature_importance}
\end{table}

Notably, \textbf{Positional Encoding} emerges as the most influential feature, indicating that the model heavily relies on the structural positioning of nodes within the network. Other network-related features, such as \textbf{Capacity Centrality}, underscore the predictive significance of controlling capacity within the network's topology.  

While economic features \textbf{Fee Rate, Remote Fee, Fee Ratio, and Local Fee} exhibit lower MDI values compared to network-based features, they still demonstrate the role of cost dynamics in shaping the model’s decision-making process. \textbf{Max HTLC (\%)}, despite ranking lower, suggests that constraints on payment size contribute to predicting channel balances. These economic features highlight how fee structures and transactional limits influence fund distribution and flow patterns within the network.  

Overall, economic and network features contribute nearly equally to model importance, suggesting a balanced reliance on both structural and financial attributes. Together, these factors enhance the model’s ability to make nuanced predictions of channel balances.

\section{Discussion}
\label{sec:discussion}

This study was conducted to explore the utility of machine learning models in predicting channel balances within the Lightning Network. A novel aspect to this method is the incorporation of graph features into the machine learning models. %This approach leverages the inherent structure of the network, utilizing the relational and topological data that characterizes the connections between nodes and channels. 
The results suggest that machine learning techniques can provide accurate predictions of channel balances. %This indicates a promising direction for further research and potential practical application in enhancing the efficiency of network operations. 

While probing techniques can estimate channel balances, they often introduce significant overhead and pose potential privacy risks. In contrast, our balance interpolation approach relies solely on crowdsourced data, offering a probabilistic estimation that minimizes the need for frequent probing. This reduces operational overhead and enhances user privacy by decreasing direct channel queries.

\section{Future Work}
\label{sec:futurework}
A natural next step is to develop an enhanced pathfinding algorithm that integrates the predictive model into a Lightning Node's routing decisions. The enhanced pathfinding algorithm would identify the most reliable path from a source to a destination through a three-step process:

\begin{enumerate}
    \item \textbf{Predict} the balance of each edge in the network.
    \item \textbf{Assign} a cost to each edge based on its predicted balance.
    \item \textbf{Compute} the optimal path using \textbf{Dijkstra’s algorithm}, selecting the path with the lowest cost.
\end{enumerate}

The cost function is the \textit{negative logarithm} of the predicted balance, a technique shown in prior work to be effective for optimizing pathfinding reliability even with uncertain channel balances\cite{pickhardt2021optimally}.

\bibliographystyle{ieeetr} % We choose the "plain" reference style
\bibliography{bibliography} % Entries are in the refs.bib file
\end{document}